\documentclass[runningheads]{llncs}
\usepackage[T1]{fontenc}
\usepackage{graphicx,verbatim}

\usepackage{booktabs}
\usepackage{multirow}

\usepackage{amsfonts}
\usepackage{amssymb}
\usepackage{amsmath}
\usepackage{subfig}
\usepackage{mathtools}
\usepackage{hyperref}
\usepackage{pifont}

\begin{document}
\title{Patient-specific radiomic feature selection with reconstructed healthy persona of knee MR images}
\titlerunning{Patient-specific radiomics with persona}

%
\begin{comment}  %% Removed for anonymized MICCAI 2025 submission
\author{First Author\inst{1}\orcidID{0000-1111-2222-3333} \and
Second Author\inst{2,3}\orcidID{1111-2222-3333-4444} \and
Third Author\inst{3}\orcidID{2222--3333-4444-5555}}
%
\authorrunning{F. Author et al.}
\institute{Princeton University, Princeton NJ 08544, USA \and
Springer Heidelberg, Tiergartenstr. 17, 69121 Heidelberg, Germany
\email{lncs@springer.com}\\
\url{http://www.springer.com/gp/computer-science/lncs} \and
ABC Institute, Rupert-Karls-University Heidelberg, Heidelberg, Germany\\
\email{\{abc,lncs\}@uni-heidelberg.de}}

\end{comment}

\author{Yaxi Chen\inst{1,3}\orcidID{0009-0007-5906-899X} \and Simin Ni\inst{2}\orcidID{0009-0007-2780-6118} \and
Aleksandra Ivanova\inst{2}\orcidID{0009-0000-4113-8928} \and
Shaheer U. Saeed\inst{3,4}\orcidID{0000-0002-5004-0663} \and
Rikin Hargunani\inst{5}\orcidID{0000-0002-0953-8443} \and
Jie Huang\inst{1}\orcidID{0000-0001-7951-2217} \and
Chaozong Liu\inst{2,5}\orcidID{0000-0002-9854-4043} \and
Yipeng Hu\inst{3,4}\orcidID{0000-0003-4902-0486}}

\authorrunning{Y. Chen et al.}


\institute{Mechanical Engineering Department, University College London, London, UK \and Institute of Orthopaedic \& Musculoskeletal Science, University College London, Royal National Orthopaedic Hospital, Stanmore, UK \and UCL Hawkes Institute, University College London, London, UK \and Department of Medical Physics and Biomedical Engineering, University College London, London, UK \and
Royal National Orthopaedic Hospital, Stanmore, UK}

%\author{Anonymized Authors}  %% Added for anonymized MICCAI 2025 submission
%\authorrunning{Anonymized Author et al.}
%\institute{Anonymized Affiliations \\
%    \email{email@anonymized.com}}

\maketitle              % typeset the header of the contribution
\begin{abstract}
Classical radiomic features (e.g., entropy, energy) have been designed to describe image appearance and intensity patterns. These features are directly interpretable and readily understood by radiologists. Compared with end-to-end deep learning (DL) models, lower dimensional parametric models that use such radiomic features offer enhanced interpretability but lower comparative performance in clinical tasks. In this study, we propose an approach where a standard logistic regression model performance is substantially improved by learning to select radiomic features for individual patients, from a pool of candidate features. This approach has potentials to maintain the interpretability of such approaches while offering comparable performance to DL. In addition, we also propose to expand the feature pool by generating a patient-specific healthy persona via mask-inpainting using a denoising diffusion model trained on healthy subjects. Such a pathology-free baseline feature set allows not only further opportunity in novel feature discovery but also improved condition classification.
We demonstrate our method on multiple clinical tasks of classifying general abnormalities, anterior cruciate ligament tears, and meniscus tears. Experimental results demonstrate that our approach achieved comparable or even superior performance than state-of-the-art DL approaches while offering added interpretability through the use of radiomic features extracted from images and supplemented by generating healthy personas. Example clinical cases are discussed in-depth to demonstrate the intepretability-enabled utilities such as human-explainable feature discovery and patient-specific location/view selection. These findings highlight the potentials of the combination of subject-specific feature selection with generative models in augmenting radiomic analysis for more interpretable decision-making. The codes are available at: \url{https://github.com/YaxiiC/RadiomicsPersona.git}
\end{abstract}
\section{Introduction}

Anterior cruciate ligament (ACL) tears are the most frequent acute knee injuries\cite{salzler2015state}. Diagnosis is primarily based on physical examinations for knee instability. For instance, The clinical Lachman test is reported to be 81\% sensitive and 81\% specific for diagnosing a complete ACL rupture~\cite{van2013methods}. Meniscus injuries are a common type of damage to cartilage in the knee, arising either from acute trauma or degenerative changes~\cite{noble1975pathology}. Physical examination tests diagnose meniscus tears with a 60--70\% sensitivity and a 70\% specificity~\cite{hegedus2007physical}. Due to low accuracy in these physical examinations, magnetic resonance (MR) imaging stands as the gold-standard for diagnoses, offering critical characterization of the tear depth, location, pattern, tissue quality, and integrity of any previous meniscal repair.

\noindent\textbf{End-to-end Deep Learning} has been employed in most existing automated approaches to MR-based diagnosis of knee injuries. Most previous studies in this area have utilized end-to-end convolutional neural networks (CNNs). For instance, Bien et al.~\cite{bien2018deep} utilized an end-to-end classifier and Tsai et al.~\cite{tsai2020knee} applied an a lightweight efficiently layered network (ELNet). While methods like class activation mappings~\cite{bien2018deep} have been used to enhance interpretability, a major limitation of these DL models remains their “black box” nature. The lack of transparency and thus human understanding of how image-level features are learned and decisions are made remains an interesting challenge.

\noindent\textbf{Radiomics} has long contributed to screening, diagnosis, treatment planning, and follow-up~\cite{zhang2023radiomics}. 
%By extracting quantitative features from medical images in a reproducible manner, radiomics enables the development of robust predictive models that correlate these features with various clinical outcomes. 
%Radiomics features can be broadly categorized into hand-crafted features () and DL features obtained through CNNs. 
Hand-crafted radiomics features, e.g. intensity, shape, texture, wavelets, provide specific information regarding region of interest (ROI), while DL features may capture more complex patterns~\cite{soffer2019convolutional}. However, these DL-based radiomics also inherits the interpretability challenge common to most DL methods. Integrating hand-crafted radiomics features and DL can potentially yield more precise and robust prognostic models~\cite{fan2024automated}. % Nonetheless, interpretability remains a key concern for clinical acceptance, motivating ongoing research into explainable and transparent radiomics pipelines.

\noindent\textbf{Generative models} have been extensively applied to medical imaging for tasks such as image quality enhancement, domain transfer, and augmentation of training data \cite{pmlr-v227-saeed24a,fu2023recycling}. For instance, generative adversarial networks (GANs) \cite{goodfellow2014generative} were employed to reconstruct images from incomplete data and reduce noise in low-dose CT and under-sampled MRI scans. Recently, denoising diffusion probabilistic model (DDPM) \cite{ho2020denoising,pmlr-v139-nichol21a} advanced image generation tasks including those from medical imaging~\cite{kather2022medical}. Generative models have also been applied in anomaly detection and image-to-image translation. For example a localized reconstruction approach used a single-mask inpainting technique to refine both anomaly detection and segmentation outcomes~\cite{colussi2024loris}. %Generative models can aid clinical automation by generating images that enhance AI-driven diagnostics.
Besides, exploring generative models in our diagnostic tasks and its interpretation remains limited.

In this work, we propose a DL-based radiomic feature selection framework for musculoskeletal disease classification, incorporating a healthy persona - a synthesized patient-specific healthy baseline for pathological regions. The term persona, inspired by the namesake concept in psychology \cite{jung2014two}, represents a predicted healthy counterpart of a patient’s anatomy without the imaged manifestation of the pathology of interest. Our approach employs a feature-weighting neural network to predict probabilities of the feature being selected, dynamically adjusting the weighting/selection of radiomic features extracted from both the pathological MRI and its healthy persona. These selected features serve as inputs to a logistic regression model, which performs a downstream clinical task. The feature-weighting neural network and logistic regression function are trained simultaneously supervised solely by the downstream task labels. To construct the healthy persona, we utilize a DDPM trained exclusively on healthy MRI scans, to reconstruct a pathology-free version of the ROI, enabling direct comparisons between the original pathology and its synthetic healthy baseline learned from healthy population. By incorporating this patient-specific persona and radiomic feature selection, this framework provides a more interpretable and clinically improved assessment of disease progression.

To summarize, our main contributions are: 1) We introduce a framework that learns individualized weighting/ selection for patch-based radiomic features, offering more precise disease characterization; 2) We propose a method to reconstruct a healthy persona for each patient’s pathological image to augment radiomic features that are subsequently used to characterize diseases; 3) We achieved comparable or even superior performance compared to traditional end-to-end approaches in knee joint MRI analysis, while offering enhanced interpretability, flexibility, and revealing potential imaging biomarkers; 4) All code is openly shared to ensure reproducibility and advance radiomics-based medical image analysis.

\section{Methods}
\label{sec:method}

%%%%%%%%%%%%%%%%%%%%%%%%%%%%%%%%%%%%%%
%%%%%%%% METHOD - SHAHEER %%%%%%%%%%%%

%Our proposed framework first computes radiomic features from an image or an image patch that contains a ROI, before a neural network predicts the weights for, or selects, features that are used as input to a logistic regression model. The model predicts disease classification. The feature weighting/ selection is learnt implicitly when the disease classification is optimized, and due to the use of a neural network to predict feature weights/ selections, the framework can dynamically change weights/ selections for each image, making our framework patient-specific.

% g(x; \theta) = P(f | x ; \theta)

% pool of features

\noindent{\textbf{Pool of features:}}
A feature-extraction function $e(\mathbf{x})=\mathbf{f} = \{f_i\}_{i=1}^F$ computes a fixed pool of $F$ features $\mathbf{f} = \{f_i\}_{i=1}^F$, from an image ROI patch $\mathbf{x}\in\mathbb{R}^{H\times W\times D}$, which corresponds to the location of the pathology, where $H\times W\times D$ denote the height, width, and depth of the patch, respectively.

% feature search
\noindent{\textbf{Feature probability prediction:} A feature-weighting neural network predicts feature probabilities:

$$g_{\theta}(\mathbf{x}) = \mathbf{p} = \{p_i\}_{i=1}^F = P(\mathbf{f} | \mathbf{x}; \theta),$$

\noindent where $\mathbf{p} = \{p_i\}_{i=1}^F = P(\mathbf{f} | \mathbf{x}; \theta)$ denotes the feature probabilities given an image patch $\mathbf{x}$, and $\theta$ denotes neural network parameters.

\noindent{\textbf{Downstream task:} The downstream task uses logistic regression to classify an image into categories (such as classifying an image into disease categories). The model inputs are weighted features
$\mathbf{f}^w = \{p_i ~f_i\}_{i=1}^F$, which are used in the logistic regression model to predict downstream task class probabilities $r_\phi(\mathbf{f}^w) = P(\mathbf{c} | \mathbf{x}, \mathbf{p}; \theta, \phi),$ where the downstream task classification probabilities, given the image $\mathbf{x}$ and feature probabilities $\mathbf{p}$, are denoted by $P(\mathbf{c} | \mathbf{x}, \mathbf{p}; \phi)$, where the image features are computed from the image patch $e(\mathbf{x})=\mathbf{f}$ and the feature probabilities are computed using the feature selection network $g_{\theta}(\mathbf{x})=\mathbf{p}$, with parameters $\theta$. The classifier parameters are denoted by $\phi$.

\noindent{\textbf{Probabilistic interpretation:} Using the chain rule of probability, the joint probability of the classes and features, given an image can be expressed as:

$$
P(\mathbf{c}, \mathbf{f} | \mathbf{x}; \theta, \phi) = P(\mathbf{c} | \mathbf{f}, \mathbf{x}; \phi) \times P(\mathbf{f}| \mathbf{x}; \theta),
$$

\noindent where $P(\mathbf{f} | \mathbf{x}; \theta)$ is predicted by the feature selection network $g_{\theta}(\mathbf{x})$, and $P(\mathbf{c} | \mathbf{f}, \mathbf{x}; \phi)$ is predicted by the logistic regression classifier $r_{\phi}(\mathbf{f}^w)$.

The goal during training is to model the classification probability given the image, $P(\mathbf{c} | \mathbf{x}; \theta, \phi)$, where feature probabilities act as intermediate variables and we do not need to explicitly model $P(\mathbf{c}, \mathbf{f} | \mathbf{x}; \theta, \phi)$. Since the feature extraction function $\mathbf{f}= e(\mathbf{x})$ is deterministic, the conditional distribution $P(\mathbf{f}|\mathbf{x}; \theta, \phi)$ is concentrated at $\mathbf{f}=e(\mathbf{x})$ (behaving like a Dirac-delta function with probability 1 at $\mathbf{f}= e(\mathbf{x})$). This allows simplification to marginalise out the intermediate variables in two steps: 1) simplify the expression using the chain rule, and 2) collapse the integral due to the deterministc function $\mathbf{f}= e(\mathbf{x})$. This gives:

\begin{align*}
    \int P(\mathbf{c}, \mathbf{f} | \mathbf{x}; \theta, \phi) d\mathbf{f}, &=  \int P(\mathbf{c}|\mathbf{f}, \mathbf{x}; \phi) ~ P(\mathbf{f}| \mathbf{x}; \theta) d\mathbf{f},\\
    &= P(\mathbf{c} | \mathbf{f}= e(\mathbf{x}), \mathbf{x}; \theta, \phi),\\
     &= P(\mathbf{c} | \mathbf{x}; \theta, \phi) 
\end{align*}

\noindent{\textbf{Training and optimization:} Following from the simplification, given a dataset of $N$ training examples ${(\mathbf{x}^{i}, \hat{\mathbf{c}}^{i})}_{i=1}^N$, where $\hat{\mathbf{c}}^{i}$ is the ground-truth label for image $\mathbf{x}^{i}$, the objective is to find parameters $(\theta, \phi)$ that maximise the likelihood of predicting the correct labels:
$\mathcal{L}(\theta, \phi) = \prod_{i=1}^N P(\hat{\mathbf{c}}^i | \mathbf{x}^i; \theta, \phi).$
The negative log-likelihood is used to facilitate the commonly-adopted cross-entropy loss in classification tasks:

$$
\mathcal{L}_{loss}(\theta, \phi) = - \sum_{i=1}^N \log P(\hat{\mathbf{c}}^i | \mathbf{x}^i; \theta, \phi)
$$
Minimizing this loss using gradient descent ensures that the predicted class probabilities align with the ground-truth labels, i.e. 
$(\theta^*, \phi^*) = \arg \min_{\theta, \phi} \mathcal{L}_{loss}(\theta, \phi).$

In summary, the likelihood function captures how well the model's predicted probabilities align with the ground-truth labels, and minimising loss directly trains the model to improve classification accuracy as well as predict intermediate feature weightings/ probabilities that ought to aid this classification.

\noindent{\textbf{Details of feature extraction:} The feature extractor $e(\mathbf{x})$ first divides the image $\mathbf{x}$ into $P$ subpatches.
% , denoted as $\{p_i\}_{i=1}^P$. 
For each subpatch, $m$ radiomic features are computed in three spatial dimensions, giving a total of $3mP$ features, i.e. $F=3mP$.

\noindent{\textbf{Expanding the pool of features through a healthy persona:} %In this work, we construct a healthy persona $\mathbf{x^\textit{persona}}$ for each image and add this to our pool of features. 
First, we train a 3D conditional DDPM model \cite{ho2020denoising} using healthy patient images. The model is trained to reconstruct healthy images from a masked version of the original images. The masks are applied to the center of the image, where all pathological features are considered to be present. At inference, we can then mask a pathological image and the DDPM can then reconstruct the image, leading to its healthy persona $\mathbf{x^\textit{persona}}$. Further details on datasets and DDPM model are outlined in Sec.\ref{subsec:dataset}. For each image, $3mP$ radiomic features are then computed on the healthy persona reconstructed by the DDPM and added to the pool of features, making the total number of features per-image $F=6mP$.

\begin{figure}
    \centering
     \includegraphics[width=0.85\textwidth]{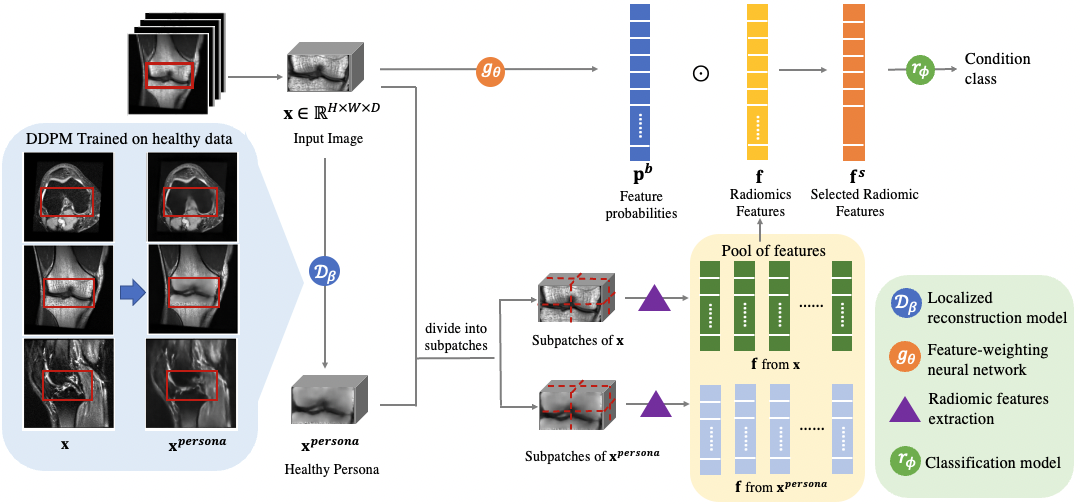}
    \caption{Overview of the radiomic feature selection framework (details in Sec.\ref{sec:method}).} %A DDPM reconstructs a healthy persona from the pathological ROI. Radiomic features are extracted from both the original and persona images, forming a feature pool. A feature-weighting neural network predicts feature relevance $\mathbf{p}$, and the selected features $\mathbf{f}^w$ are used in a logistic regression classifier for disease prediction.}
    \label{fig1}
\end{figure}

\noindent{\textbf{Feature selection at inference:} At inference, the predicted feature probabilities $g_{\theta}(\mathbf{x}) = \mathbf{p} = \{p_i\}_{i=1}^F$ are binarized to conduct a hard selection as opposed to the soft weighting during training. The binarized feature probabilities are 
$\mathbf{p}^b = \{p^b_i\}_{i=1}^F$, where ${p}_i^b$ = 1 if $p_i \geq T$, 0 otherwise, \({T} \) is an application-specific threshold hyperparameter. These binarized values are then used to select features, resulting in the final set of selected features $
\mathbf{f}^s = \{p^b_i ~f_i\}_{i=1}^F,
$ which can be used to determine the final predicted disease classification from logistic regression: $
r_\phi(\mathbf{f}^s)
$.

\noindent Note that the expectation of the feature weighting during training is proportional to the selected feature at inference, $\mathbb{E} \left[\mathbf{f}^w  \right] \propto \mathbf{f}^s$ and $P(\mathbf{f}^w | \mathbf{x}; \theta) \propto P(\mathbf{f}^s | \mathbf{x}; \theta)$.

\section{Experiments and Results}

\subsection{Dataset and Implementation Details}

\label{subsec:dataset}
\textbf{Dataset and Preprocessing:} We use the MRNet dataset~\cite{bien2018deep}, which consists of 1,370 3D knee MRI studies, acquired from axial, coronal, and sagittal views. Each case is labeled for the presence of abnormalities, an ACL tear, or a meniscal tear. For our study, we utilize 1,130 cases for training and 120 for validation. All images are preprocessed and resized to \( 32 \times 128 \times 128 \), tested with and without an affine registration (NiftyReg~\cite{modat2010fast} in this study).

\noindent\textbf{DDPM model training:}
We implement a DDPM3D to reconstruct MRI scans by progressively denoizing corrupted inputs. Our model follows a diffusion process parameterized by a sequence of variance schedules \(\beta_t\), where we define \(\beta_t\) to linearly increase from \(10^{-4}\) to \(0.02\) over \(T=1000\) timesteps. %The forward process gradually corrupts an input \( x_0 \) into a Gaussian noise distribution, while the reverse process, learned through a 3D UNet, estimates the noise at each timestep. 
During training, we utilize the healthy subset of the MRNet dataset, which consists of MRI volumes labeled as \(abnormal=0\), comprising 218 patients and a total of 654 scans across three views. Each scan was masked by a central bounding box, with a \(50\%\) depth, a \(30\%\) height, and a \(50\%\) width of the original image size. %The DDPM3D model is trained to reconstruct the missing structure by generating the corresponding healthy \textit{persona}. %The reconstruction loss is computed using the mean squared error (MSE) between \( I_{\text{ROI}}^{\text{persona}} \) and the original uncorrupted region, ensuring anatomically plausible reconstructions.
%\noindent\ By training exclusively on healthy MRI scans, \( \mathcal{D}_{\theta} \) effectively learns to inpaint pathological ROIs with anatomically realistic structures, facilitating robust reconstruction for downstream analysis.

\noindent\textbf{Feature Extraction and selection:}  
We extract features using a PyTorch implementation based on the PyRadiomics library~\cite{van2017computational}. We compute \textit{First-Order Features} and \textit{3D Shape Features}, resulting in a total of 35 feature types. Both the pathological image and its DDPM-generated healthy persona are divided into \(2 \times 2 \times 2\) subpatches. For each patient, we extract features from these subpatches in the axial, coronal, and sagittal planes, resulting in a final feature vector of size \(3 \times 8 \times 2 \times (35 + 3) = 1824\). 
%In this notation, \(3\) denotes the number of MRI views, \(8\) is the total number of subpatches per view, \(2\) captures both the pathological image and healthy (\textit{persona}) images, and \(35 + 3\) corresponds to the dimension of the feature vector (35 radiomic features plus 3 spatial coordinates). 
We employ a 3DResNet-18~\cite{he2016deep}, as the feature-weighting neural network to learn patient-specific feature probabilities.

%\noindent\textbf{Classification pipeline:}
%To classify pathological conditions, we employ a multi-stage learning framework that integrates patient-specific feature weighting and logistic regression. The classifier, denoted as \( C_{\beta} \), takes the weighted feature vector \( \mathbf{F}_c \) as input and outputs the probability of each pathology class.

%\noindent\textbf{Feature selection and weighting:}
%We employ a ResNet-based feature selector that learns patient-specific feature weights. we used 3D ResNet-18~\cite{he2016deep} adapted for volumetric medical images. 

%The output of $$g_{\theta}$$ is a patient-specific probability vector, which is used to modulate the extracted radiomic features via element-wise multiplication.

\noindent\textbf{Classification model:}
During training, the probability-weighted feature vector \(\mathbf{f}^w\) is fed into the logistic regression classifer to predict class probabilites for the downstream classification task. After training, the optimal classification thresholds are determined using Youden’s index on the validation set. During inference, we apply a threshold 0.4 for hard feature selection. We also examined the effect of varying the feature selection threshold. As expected, higher thresholds resulted in poorer performance (further details in Tab.~\ref{tab1}), and 0.4 was selected for its balance between performance and number of selected features.

\begin{table}[!ht]
\renewcommand{\thetable}{1}
\centering
\caption{Comparison of different configurations and methods. N: Number of subpatches; PFS: Patient-specific features selection; HP: Healthy persona; T: Feature selection threshold; Reg: Registration in preprocessing. %The evaluation metrics include Acc (Accuracy), Sen (Sensitivity), Spe (Specificity), and AUC (Area Under the Curve).
* denotes reproduced results}
\label{tab1}

\renewcommand{\arraystretch}{0.9} % Reduce row spacing

\setlength{\tabcolsep}{1.3pt}
\fontsize{8pt}{9pt}\selectfont

%\begin{tabular}{|c|cccccc|c|cccc|}
\begin{tabular}{|@{\hskip 0.9pt}c|@{\hskip 0.9pt}c|@{\hskip 0.9pt}c|@{\hskip 0.9pt}c|@{\hskip 0.9pt}c|@{\hskip 0.9pt}c|@{\extracolsep{1.5pt}}c|c@{\hskip 1.5pt}c@{\hskip 1.5pt}c@{\hskip 1.5pt}c|}

\hline
 & & & & & & & \multicolumn{4}{c|}{\textbf{Evaluation Metrics}} \\ 
\cline{8-11} 
\textbf{Method} & \textbf{N} & \textbf{PFS} & \textbf{HP} & \textbf{T} & \textbf{Reg} & \textbf{Type} & \textbf{Acc} & \textbf{Sen} & \textbf{Spe} & \textbf{AUC} \\ 
\hline

\hline
\textbf{\multirow{3}{*}{Ours}} &\multirow{3}{*}{2*2*2} & \multirow{3}{*}{\ding{51}} & \multirow{3}{*}{\ding{51}} & \multirow{3}{*}{0.4} & \multirow{3}{*}{\ding{51}} 
& abn & \textbf{0.90±0.13} & 0.94±0.10 & 0.77±0.25 & 0.85±0.16 \\
&&&&&& acl & 0.81±0.10 & 0.92±0.08 & 0.66±0.23 & 0.80±0.12 \\
&&&&&& men & 0.82±0.11 & 0.81±0.14 & 0.88±0.14 & 0.84±0.11 \\
\hline

\textbf{\multirow{3}{*}{Ours}} & \multirow{3}{*}{3*3*3} & \multirow{3}{*}{\ding{51}} & \multirow{3}{*}{\ding{51}} & \multirow{3}{*}{0.4} & \multirow{3}{*}{\ding{51}} 
& abn & 0.84±0.12 & 0.96±0.08 & 0.54±0.33 & 0.75±0.25 \\
&&&&&& acl & \textbf{0.87±0.16} & 0.88±0.13 & 0.87±0.22 & 0.87±0.15 \\
&&&&&& men & \textbf{0.87±0.23} & 0.88±0.16 & 0.87±0.30 & 0.88±0.20 \\
\hline
\multirow{3}{*}{FST=0.0} & \multirow{3}{*}{2*2*2} & \multirow{3}{*}{\ding{51}} & \multirow{3}{*}{\ding{51}} & \multirow{3}{*}{0.0} & \multirow{3}{*}{\ding{51}} 
& abn & \textbf{0.90±0.11} & 0.99±0.03 & 0.63±0.42 & 0.81±0.21 \\
&&&&&& acl & 0.84±0.13 & 0.83±0.23 & 0.84±0.16 & 0.83±0.15 \\
&&&&&& men & \textbf{0.83±0.16} & 0.82±0.26 & 0.83±0.13 & 0.83±0.17 \\
\hline
\multirow{3}{*}{FST=0.5} & \multirow{3}{*}{2*2*2} & \multirow{3}{*}{\ding{51}} & \multirow{3}{*}{\ding{51}} & \multirow{3}{*}{0.5} & \multirow{3}{*}{\ding{51}} 
& abn & 0.76±0.18 & 0.72±0.22 & 0.90±0.15 & 0.81±0.17 \\
&&&&&& acl & 0.75±0.17 & 0.79±0.18 & 0.71±0.36 & 0.75±0.15 \\
&&&&&& men & 0.78±0.13 & 0.74±0.11 & 0.85±0.17 & 0.82±0.09 \\
\hline
\multirow{3}{*}{1patch}& \multirow{3}{*}{1*1*1} & \multirow{3}{*}{\ding{51}} & \multirow{3}{*}{\ding{51}} & \multirow{3}{*}{0.4} & \multirow{3}{*}{\ding{51}} 
& abn & 0.70±0.35 & 0.64±0.46 & 0.90±0.15 & 0.78±0.23 \\
&&&&&& acl & 0.73±0.15 & 0.96±0.06 & 0.56±0.26 & 0.76±0.13 \\
&&&&&& men & 0.60±0.08 & 0.76±0.27 & 0.50±0.28 & 0.64±0.05 \\
\hline
\multirow{3}{*}{NoPsa}& \multirow{3}{*}{2*2*2} & \multirow{3}{*}{\ding{51}} & \multirow{3}{*}{\ding{55}} & \multirow{3}{*}{0.4} & \multirow{3}{*}{\ding{51}} 
& abn & 0.77±0.13 & 0.75±0.19 & 0.87±0.18 & 0.81±0.10 \\
&&&&&& acl & 0.66±0.18 & 0.92±0.10 & 0.45±0.36 & 0.68±0.15 \\
&&&&&& men & 0.72±0.10 & 0.80±0.15 & 0.68±0.10 & 0.74±0.10 \\
\hline
\multirow{3}{*}{NoFS}& \multirow{3}{*}{2*2*2} & \multirow{3}{*}{\ding{55}} & \multirow{3}{*}{\ding{51}} & \multirow{3}{*}{0.4} & \multirow{3}{*}{\ding{51}} 
& abn & 0.81±0.03 & 0.93±0.05 & 0.53±0.22 & 0.68±0.09 \\
&&&&&& acl & 0.77±0.11 & 0.83±0.12 & 0.60±0.32 & 0.72±0.14 \\
&&&&&& men & 0.71±0.05 & 0.78±0.07 & 0.66±0.11 & 0.72±0.03 \\
\hline
%\multirow{3}{*}{1view}& \multirow{3}{*}{2*2*2} & \multirow{3}{*}{Y} & \multirow{3}{*}{Y} & \multirow{3}{*}{h} & \multirow{3}{*}{1} & \multirow{3}{*}{Y} 
%& abn & 0.86 & 0.98 & 0.42 & 0.70 \\
%&&&&&&& acl & \textbf{0.92} & 0.93 & 0.91 & 0.92 \\
%&&&&&&& men & 0.78 & 0.96 & 0.63 & 0.80 \\
%\hline
\multirow{3}{*}{NoReg}& \multirow{3}{*}{2*2*2} & \multirow{3}{*}{\ding{51}} & \multirow{3}{*}{\ding{51}} & \multirow{3}{*}{0.4} & \multirow{3}{*}{\ding{55}} 
& abn & 0.77±0.08 & 0.57±0.13 & 0.96±0.04 & 0.76±0.06 \\
&&&&&& acl & 0.73±0.10 & 0.40±0.29 & 0.99±0.03 & 0.70±0.10 \\
&&&&&& men & 0.67±0.15 & 0.99±0.03 & 0.41±0.27 & 0.70±0.14 \\
\hline
\multirow{3}{*}{\textbf{ELNet*}} & \multirow{3}{*}{-} & \multirow{3}{*}{-} & \multirow{3}{*}{-} & \multirow{3}{*}{-} & \multirow{3}{*}{-} 
& abn & 0.80±0.01 & 0.95±0.01 & 0.22±0.03 & 0.73±0.01 \\
&&&&&& acl & 0.70±0.10 & 0.72±0.26 & 0.68±0.04 & 0.73±0.08 \\
&&&&&& men & 0.65±0.06 & 0.61±0.13 & 0.63±0.07 & 0.69±0.03 \\
\hline
\multirow{3}{*}{\textbf{MRNet}} & \multirow{3}{*}{-} & \multirow{3}{*}{-} & \multirow{3}{*}{-} & \multirow{3}{*}{-} & \multirow{3}{*}{-} 
& abn & 0.85 & 0.87 & 0.71 & 0.94 \\
&&&&&& acl & 0.87 & 0.76 & 0.97 & 0.92 \\
&&&&&& men & 0.73 & 0.71 & 0.74 & 0.82 \\

\hline

\multirow{3}{*}{\textbf{SKID}} & \multirow{3}{*}{-} & \multirow{3}{*}{-} & \multirow{3}{*}{-} & \multirow{3}{*}{-} & \multirow{3}{*}{-} 
& abn & 0.87 & 0.98 & 0.49 & 0.90 \\
&&&&&& acl & 0.80 & 0.74 & 0.85 & 0.89 \\
&&&&&& men & 0.73 & 0.92 & 0.57 & 0.81 \\
\hline

\end{tabular}
\end{table}

\noindent\textbf{Comparison and Ablation Experiments:}  In summary, we compare our method with end-to-end approaches, including MRNet~\cite{bien2018deep}, ELNet~\cite{tsai2020knee}, and SKID~\cite{manna2023self} without pre-training, achieving competitive performance with much enhanced interpretability. To evaluate the impact of key components, we conducted ablation studies. Removing the feature-weighting neural network and using raw radiomic features resulted in suboptimal performance, highlighting the importance of adaptive weighting. Different number of subpatch (\(1 \times 1 \times 1\) and \(3 \times 3 \times 3\)) were analyzed, with \(2 \times 2 \times 2\) providing the best balance of performance and interpretability. Excluding persona and using only the original ROI reduced classification accuracy, demonstrating the benefit of synthetic healthy baselines. Omitting registration led to decreased performance.

\subsection{Results and Interpretation}
\label{sec:results}

Our proposed method with \(2 \times 2 \times 2\) subpatches outperforms existing models in detecting abnormalities and meniscal tears, achieving 0.90 and 0.82 accuracy, respectively. This is significantly better than ELNet, an end-to-end DL method, in meniscus tear detection (all \textit{p-values}$<$0.050 across all evaluated metrics, paired t-test at a significance level $\alpha$=0.05). Increasing the number of subpatches to \( 3 \times 3 \times 3 \) further enhances performance, achieving an accuracy of 0.87 for both ACL and meniscal tear detection. This surpasses the reported accuracy of MRNet and SKID (data unavailable for paired tests). Our method exhibits high sensitivity while maintaining comparable specificity, demonstrating comparable or even superior performance across multiple configurations. 

\begin{figure}
    \centering
     \includegraphics[width=0.85\textwidth]{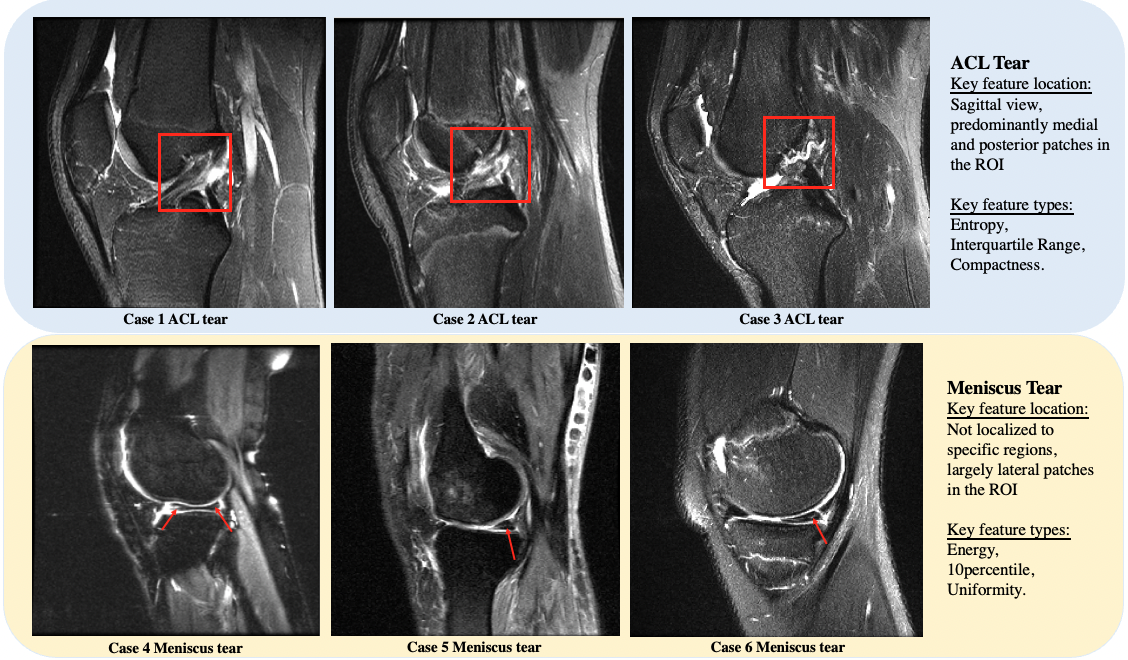}
    \caption{Representative cases of ACL and Meniscus Tears (details in Sec.\ref{sec:results}).}%For ACL tears (top row), key features are localized in the sagittal view, primarily in the medial and posterior regions of the ROI, with high \emph{Entropy}, \emph{Interquartile Range}, and \emph{Compactness}. For Meniscus tears (bottom row), key features are more diffuse, with lateral patches showing significant variations in \emph{Energy}, \emph{10th percentile intensity}, and \emph{Uniformity}. }
    \label{fig2}
\end{figure}

Ablation studies demonstrate that our method significantly outperforms the configurations without registration and without persona in ACL detection, achieving higher accuracy with \textit{p-values} of 0.018 and 0.023, respectively. Furthermore, it achieves a statistically significant improvement in accuracy over the single-patch configuration in meniscus detection (\textit{p-values}=0.043).

\noindent
\textbf{ACL tears case studies:} 
For ACL injuries, key features are concentrated in the sagittal view and the medial/posterior regions of the ROI. This aligns with both clinical expectations, as the sagittal view provides the clearest visualization of ACL continuity, and anatomically, the ACL originates from the intercondylar eminence of the tibia and inserts onto the medial aspect of the lateral femoral condyle\cite{petersen2007anatomy}. The key radiomic features in the medial and posterior subpatches for ACL tears include \emph{Entropy}, \emph{Interquartile Range}, and \emph{Compactness}. Entropy quantifies the complexity and irregularity in the image intensity patterns, typically increasing when torn ligament fibers appear irregular and heterogeneous (see Fig.~\ref{fig2}). Interquartile Range, representing the spread of voxel intensities in the middle half of the patch, captures the intensity heterogeneity indicative of discontinuous fibers. Compactness, quantifying how closely a shape approximates a sphere, reflects structural deformation and may highlight secondary signs such as posterior cruciate ligament laxity and anterior tibial translation.

\noindent 
\textbf{Meniscus tears case studies:} In contrast, key features for meniscus tear are not localized to specific regions, as lesions can occur in either or both menisci. On PD-weighted MRI, these tears characteristically appear as a linear high-intensity signal coursing through the meniscus, typically extending to its superior or inferior surface. Key radiomic features in the lateral subpatches include \emph{Energy}, which measures overall tissue integrity and decreases with disruption of the meniscal substance, \emph{Uniformity}, which detects irregular voxel intensities in torn regions, and the \emph{10percentile} feature, %the intensity value at the 10th percentile of the voxel intensity distribution,
which highlights subtle changes in the lower-intensity range of the meniscal signal. Given the variability in tear location and morphology, patient-specific radiomic feature selection remains essential for robust detection and quantification of meniscal pathology.

\noindent
\textbf{Model prediction-informed image view selection:} Building on the observation in the model prediction, and expected radiologically, that ACL-related features concentrate in the sagittal view, we refined our framework by extracting features exclusively from sagittal slices while maintaining other configurations. This approach improved the performance to 0.92 accuracy, 0.93 sensitivity, 0.91 specificity, and 0.92 AUC.

\section{Discussion and Conclusion}

Our study presents a novel framework for knee MRI analysis that integrates generative modeling and logistic regression classification to enable an individualized learning-based radiomic selection, for pathology detection with improved interpretability. By introducing a patient-specific healthy persona, we provide a baseline comparison that facilitates new insights and clinically relevant analysis of pathological regions. %The proposed approach successfully balances performance and explainability, making it a viable candidate for real-world clinical applications.
Our results demonstrate improvements over conventional end-to-end learning approaches, while maintaining an interpretable feature selection process, is a key advantage of our method. 
\section{Acknowledgement}
This work was supported by the International Alliance for Cancer Early Detection, an alliance between Cancer Research UK [EDDAPA-2024/100014] \& [C73666/A31378], Canary Center at Stanford University, the University of Cambridge, OHSU Knight Cancer Institute, University College London and the University of Manchester; and the National Institute for Health Research University College London Hospitals Biomedical Research Centre.


\bibliographystyle{splncs04}
\bibliography{MICCAI2025_paper_template.bib}

\end{document}